\documentclass[sigconf]{acmart}

\copyrightyear{2024}
\acmYear{2024}
\setcopyright{acmlicensed}
\acmConference[WWW '24 Companion] {Companion Proceedings of the ACM Web Conference 2024}{May 13--17, 2024}{Singapore, Singapore.}
\acmBooktitle{Companion Proceedings of the ACM Web Conference 2024 (WWW '24 Companion), May 13--17, 2024, Singapore, Singapore}
\acmISBN{979-8-4007-0172-6/24/05}
\acmDOI{10.1145/3589335.3651460}

\usepackage{graphicx}
\usepackage{amsmath}
\usepackage{booktabs}
\usepackage{epsfig}
\usepackage{enumitem}
\setlist[itemize]{leftmargin=3mm}
\usepackage{multirow}
\usepackage{makecell}
\usepackage{xr}
\usepackage{dsfont}
\usepackage{tcolorbox}
\usepackage[ruled,vlined,linesnumbered]{algorithm2e}

\def\eg{\emph{e.g.,}\xspace}

\def\etal{\emph{et al.}\xspace}

\begin{document}

\author{Haoxin Xu}\authornote{Both authors contributed equally to this research.}\authornote{This work was done when Haoxin was an intern at Ping An Technology.}
\affiliation{
  \institution{Harbin Engineering University}
  \country{China}}
  \email{haoxin@hrbeu.edu.cn}

\author{Zezheng Zhao}\authornotemark[1]
\affiliation{
  \institution{Ping An Technology}
  \country{China}}
  \email{zhaozezheng995@pingan.com.cn}

\author{Yuxin Cao}
\affiliation{
  \institution{Tsinghua University}
  \country{China}}
  \email{caoyx21@mails.tsinghua.edu.cn}

\author{Chunyu Chen}\authornote{Corresponding author.}
\affiliation{
  \institution{Harbin Engineering University}
  \country{China}}
  \email{springrain@hrbeu.edu.cn}
  
\author{Hao Ge}
\affiliation{
  \institution{Ping An Technology}
  \country{China}}
  \email{gehao692@pingan.com.cn}

\author{Ziyao Liu}
\affiliation{
  \institution{Nanyang Technological University}
  \country{Singapore}}
  \email{liuziyao@ntu.edu.sg}

\renewcommand{\shortauthors}{Haoxin Xu et al.}

\title{3D Face Reconstruction Using A Spectral-Based Graph Convolution Encoder}

\begin{abstract}
Monocular 3D face reconstruction plays a crucial role in avatar generation, with significant demand in web-related applications such as generating virtual financial advisors in FinTech. Current reconstruction methods predominantly rely on deep learning techniques and employ 2D self-supervision as a means to guide model learning. However, these methods encounter challenges in capturing the comprehensive 3D structural information of the face due to the utilization of 2D images for model training purposes. To overcome this limitation and enhance the reconstruction of 3D structural features, we propose an innovative approach that integrates existing 2D features with 3D features to guide the model learning process. Specifically, we introduce the 3D-ID Loss, which leverages the high-dimensional structure features extracted from a Spectral-Based Graph Convolution Encoder applied to the facial mesh. This approach surpasses the sole reliance on the 3D information provided by the facial mesh vertices coordinates. Our model is trained using 2D-3D data pairs from a combination of datasets and achieves state-of-the-art performance on the NoW benchmark. 
\end{abstract}

\begin{CCSXML}
<ccs2012>
   <concept>
       <concept_id>10010147.10010178.10010224.10010245.10010254</concept_id>
       <concept_desc>Computing methodologies~Reconstruction</concept_desc>
       <concept_significance>500</concept_significance>
       </concept>
 </ccs2012>
\end{CCSXML}

\ccsdesc[500]{Computing methodologies~Reconstruction}

\keywords{3D face reconstruction, avatar generation, single monocular image}

\maketitle
\thispagestyle{empty}
\section{Introduction}
As a pivotal research topic in the field of avatar generation, monocular face reconstruction holds considerable potential for various web applications such as generating virtual financial advisors and receptionists in FinTech, since a fast and accurate 3D face reconstruction method can save manpower costs and enhance user experience. Moreover, achieving high-precision 3D face reconstruction is imperative for facial biometric payment systems. 

Learning how to reconstruct 3D faces from images is an ill-posed inverse problem~\cite{zielonka2022towards}. Particularly, when dealing with monocular images, it becomes challenging to obtain complete 3D facial features. Recently, the seminal 3D Morphable Model (3DMM)~\cite{blanz2023morphable} brought about a significant transformation in the domain of monocular face reconstruction. The 3DMM can be regarded as a linear, low-dimensional representation of facial shape space obtained through Principal Component Analysis. By estimating the parameters of the 3DMM from an input image, the facial shape can be obtained. While the initial model-fitting problem could be addressed using optimization-based methods, recent advances have demonstrated that learning-based regression methods yield superior performance in estimating the 3DMM parameters. However, the effective use of such methods is hindered by the scarcity of large-scale 3D datasets available for training. Consequently, many existing regression-based reconstruction methods~\cite{feng2021learning, sanyal2019learning} adopt self-supervision by rendering the reconstructed mesh as 2D image. These methods rely on loss functions that evaluate the model's performance at the 2D level. Although some methods~\cite{zielonka2022towards, zhang2023accurate} have considered supervision at the 3D level, they are limited in their utilization of solely the facial mesh vertices coordinates.

To close the gap mentioned above, we propose a regression-based face reconstruction method by leveraging a Spectral-Based Graph Convolution Encoder~\cite{ranjan2018generating} to integrate both 2D and 3D structural features. Our primary objective is to extract high-dimensional structural features from the facial mesh. By combining self-supervision with 2D features and full supervision with 3D features, we aim to provide more comprehensive guidance for the model to effectively learn 3D facial features. We argue that evaluating the similarity between the reconstructed facial mesh and the target mesh should not be restricted solely to the absolute vertices positions; it is equally important to consider the relative relationships between vertices. To address this concern, we employ graph convolution techniques and utilize the graph data structure to accurately represent the relative relations and features among the vertices of the facial mesh. To combine the 2D and 3D features, we construct a dataset comprising paired 2D-3D data to train our model. The 2D image data provide self-supervision for the rendered 2D images, encompassing identity, landmarks, pixel values, and other relevant information. Meanwhile, the 3D scanning data offer comprehensive supervision for the mesh vertices and 3D structural feature vectors. All the 3D data we used adhere to a unified FLAME~\cite{li2017learning} topology, as FLAME is one of the most widely used 3DMM models. By effectively integrating these two sources of information, our model is trained within a learning-based regression framework, aiming at accurately estimating the 3DMM deformation parameters. The experimental results demonstrate the superiority of our method in the 3D face reconstruction task. Our method achieves better reconstruction results compared to previous 3D face reconstruction methods and state-of-the-art reconstruction accuracy on the NoW benchmark.

In summary, our main contributions are as follows:
\begin{itemize}
\item We propose a comprehensive pipeline that combines both 2D and 3D features to achieve accurate monocular face reconstruction. 
\item We introduce a Spectral-Based Graph Convolution Encoder that is specifically designed for 3D mesh structures. This encoder effectively extracts 3D mesh structural features and supervises the model learning process. We also propose a 3D-ID Loss, which serves as a supervision mechanism for the model to learn intricate 3D structural features of the facial mesh.
\item Our method achieves state-of-the-art reconstruction accuracy and better visualization effect on the NoW benchmark. 
\end{itemize}

\section{Related Work}
Currently, many face reconstruction methods commonly utilize Convolutional Neural Networks to extract informative features from 2D images, which are subsequently mapped to 1D vectors that represent the deformation parameters (\eg shape, expression, and pose) or the rendering parameters (\eg camera, texture, and lighting) of the 3DMM.

Due to the limited availability of 3D facial scanning data, many deep learning-based reconstruction methods have adopted self-supervised or weakly supervised training methods~\cite{feng2021learning, sanyal2019learning}. Soubhik \etal~\cite{sanyal2019learning} employed 2D landmark loss as self-supervision and integrated shape consistency as an additional supervision mechanism to ensure consistent reconstructed facial mesh shape across diverse photographs of the same individual. Subsequently, Feng \etal~\cite{feng2021learning} extended it by introducing identity and photometric losses. They further divided the monocular face reconstruction process into two stages: an initial coarse face reconstruction, followed by a subsequent refinement process that focuses on capturing intricate facial expressions and details. However, due to the reliance on projecting and rendering the predicted facial mesh back to 2D image, these methods face limitations in effectively capturing the intrinsic 3D structural features of facial models. Afterwards, 3D full supervision was additionally employed in~\cite{zielonka2022towards, zhang2023accurate}, where model training is directly supervised by utilizing the $L_1$ loss to calculate the difference between ground-truth 3D vertices and the predicted mesh vertices. These methods incorporate 3D supervision but are constrained to a solitary vertices coordinates. Therefore, they cannot capture the intricate structural features of the facial mesh in high dimensions.

\section{Method}
\begin{figure}[t]
    \centering
    \includegraphics[width=0.98\linewidth]{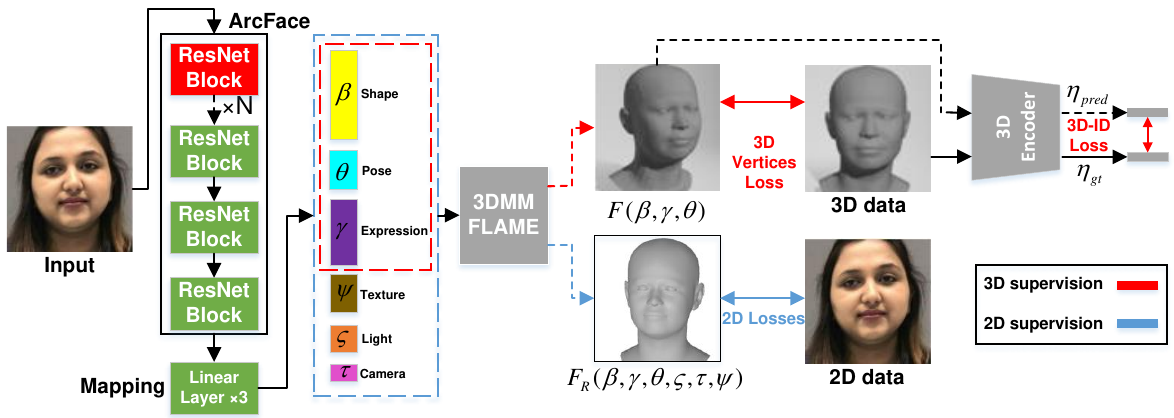}
    \vspace{-1mm}
    \caption{Overview of our method.}
    \label{fig_pipeline}
    \vspace{-5mm}
\end{figure}

\subsection{Detailed Method}
To enhance the reconstruction of facial 3D structural features, our method integrates features from both 2D and 3D levels to supervise the model learning process. The structure of our method is illustrated in Figure~\ref{fig_pipeline}. We use ArcFace~\cite{deng2020sub} as a facial ID feature extractor. Compared to traditional ResNet architectures, ArcFace demonstrates superior capabilities in effectively extracting facial image information, which can enhance the reconstruction accuracy. Following~\cite{zielonka2022towards}, we only update the parameters of the last 3 ResNet blocks of ArcFace to achieve better reconstruction generalization.

After feature extraction using ArcFace, the mapping layer produces a 1D vector with a length of 486.  This vector encompasses shape, pose, expression, texture, camera and light parameters, with dimensions of 300, 6, 100, 50, 3, and 27, respectively. These parameters govern the reconstruction of a facial mesh using the FLAME face model. Subsequently, the FLAME renderer converts the reconstructed facial mesh into a 2D image, enabling 2D self-supervision. Furthermore, the facial mesh and the ground-truth mesh undergo separate input processes into a 3D encoder, resulting in 3D structural feature vectors for each mesh. By evaluating the 3D-ID loss based on these feature vectors, our model benefits from comprehensive 3D supervision, leading to enhanced reconstruction performance. The integration of 3D supervision from the 3D encoder contributes to the overall efficacy of our approach. Additionally, the 3D vertices loss is constructed by leveraging the vertices coordinates of the mesh. This loss function directly guides the model training process by quantifying the disparity between the predicted and ground-truth vertices coordinates. Incorporating this loss enables the model to effectively incorporate and leverage valuable 3D information, thereby facilitating the learning process.

\noindent\textbf{Building 3D Encoder.}
Considering that the features of a 3D facial mesh are not represented by individual vertices but by the faces formed by connecting vertices, it becomes imperative to establish a framework that captures this relationship. In response, we propose the utilization of a graph as a suitable data structure for defining the connectivity between vertices and edges within a 3D facial mesh. Graphs are well-suited for encoding this connectivity information, facilitating a comprehensive analysis of the mesh structure. To address this, we introduce a Spectral-Based Graph Convolution Encoder on mesh. This encoder leverages spectral-based graph convolution techniques to extract high-dimensional features from the mesh, successfully operating on non-Euclidean domains. Furthermore, the extracted features derived from the graph convolution are employed to supervise the model in learning 3D-level characteristics. This supervision extends beyond the mere consideration of 3D vertices coordinates, allowing the model to grasp more nuanced and comprehensive 3D features present within the facial mesh. 

\begin{figure}[t]
    \centering
    \includegraphics[width=0.98\linewidth]{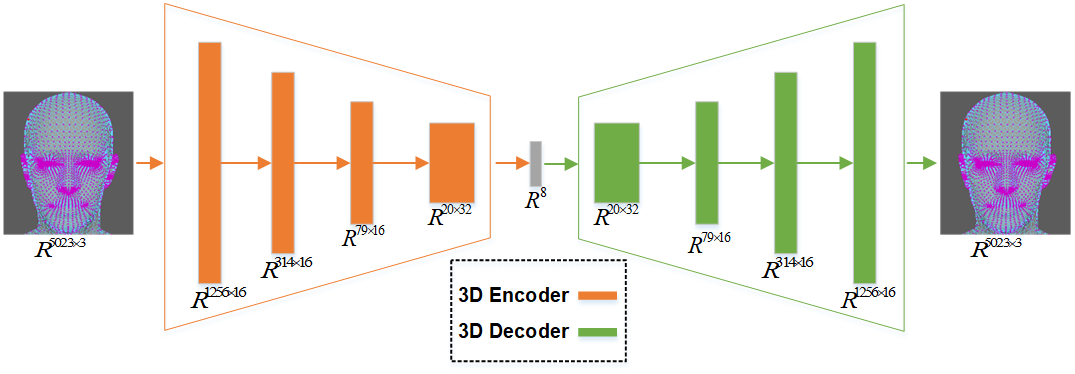}
    \vspace{-1mm}
    \caption{The structure of 3D encoder and decoder.}
    \label{fig_3Dencoder}
    \vspace{-2mm}
\end{figure}

To train such a 3D encoder, we need to simultaneously train a corresponding 3D decoder. Figure~\ref{fig_3Dencoder} illustrates the structure of 3D encoder and decoder. The encoder consists of four Chebyshev convolutional filters, each with $K = 6$ Chebyshev polynomials. A fully connected layer is used to transform the facial mesh from ${R^{5023 \times 3}}$ to an 8-dimensional feature vector. In this work, our encoder is trained using 3D scanning data with a unified FLAME topology structure. During the training process, we optimize the $L_1$ loss between the predicted mesh vertices and the ground-truth mesh vertices to guide the model's learning process. 

\subsection{Training Strategy}
Based on 2D-3D data pairs, our training process takes into account the 2D information from the input image and the 3D information from the facial mesh. The 2D data provide weak supervision in terms of photometric properties (\eg pixel values and lighting), landmark, and identity. On the other hand, the 3D data provide full supervision in terms of mesh vertices and 3D structural feature vectors. To achieve better reconstruction results, we combine both 2D and 3D supervision. Mathematically, our model training aims to minimize 
\begin{equation}
L = {\lambda _{3D}} {L_{3D}} + {\lambda _{2D}} {L_{2D}},
\label{loss_all}
\end{equation}
where ${L_{3D}}$ and ${L_{2D}}$ are 3D Loss and 2D Loss respectively, ${\lambda_{3D}}$ and ${\lambda_{2D}}$ are the weight respectively. 

\subsubsection{3D Supervision}
Since each 2D image in the training data is associated with a corresponding 3D scanned mesh, we can leverage the 3D data to provide information about the facial mesh vertices and 3D identity features. The 3D-related Loss is defined as
\begin{equation}
{L_{3D}} = {\lambda _1} {L_{vertices}} + {\lambda _2} {L_{3D - ID}},
\label{loss_3d}
\end{equation}
where ${L_{vertices}}$ and ${L_{3D - ID}}$ denote 3D Vertices Loss and 3D-ID Loss respectively, ${\lambda _1}$ and ${\lambda _2}$ represent corresponding weights.

\noindent\textbf{Vertices Loss.}
The reconstructed facial mesh $M$ can be obtained by inputting the parameters of shape $\beta$, expression $\gamma$ and pose $\theta$ to the FLAME face model $F$:
\begin{equation}
M = {F}(\beta ,\gamma ,\theta ).
\end{equation}

The learning process of our model is supervised by optimizing the $L_1$ loss between the reconstructed facial mesh and the ground-truth 3D data mesh vertices. This loss function serves as a guiding signal, directing the model towards minimizing the discrepancy between the predicted mesh and the ground-truth representation. The ${L_{vertices}}$ is defined as
\begin{equation}
{L_{vertices}} = {K_{mask}}{\left| {V - {V_{gt}}} \right|_1},
\end{equation}
where ${K_{mask}}$ is the weight of mesh vertices in different areas of the face (\eg eye, nose, forehead, mouth). ${V}$ and ${V_{gt}}$ are the vertices of the reconstructed facial mesh $M$ and the ground-truth mesh $M_{gt}$. 

\noindent\textbf{3D-ID Loss.}
To derive the structural feature vectors, we employ a 3D encoder that takes both the reconstructed facial mesh and the ground-truth mesh from the 3D data as inputs. The 3D encoder ${{\mathop{\rm E}\nolimits} _{3D}}$ effectively captures and encodes the structural characteristics of the facial meshes $M$ and $M_{gt}$ into 3D structure feature vectors $\eta_{pred}$ and $\eta_{gt}$:
\begin{equation}
\begin{array}{l}
{\eta _{pred}} = {{\rm{E}}_{3D}}(M),\\
{\eta _{gt}} = {{\rm{E}}_{3D}}({M_{gt}}).
\end{array}
\end{equation}

By optimizing the similarity between the structural feature vectors of the reconstructed facial mesh and the ground-truth 3D data mesh, we can supervise the model in reconstructing the facial 3D structural features. The ${L_{3D - ID}}$ is defined as
\begin{equation}
{L_{3D - ID}} = 1 - \frac{{{\eta _{pred}} {\eta _{gt}}}}{{{{\left\| {{\eta _{pred}}} \right\|}_2} {{\left\| {{\eta _{gt}}} \right\|}_2}}}.
\end{equation}

\subsubsection{2D Supervision}
By inputting the predicted parameters of shape $\beta$, expression $\gamma$, pose $\theta$, light $\varsigma$, camera $\tau$ and texture $\psi $ into the FLAME mesh renderer $F_R$, we can obtain the reconstructed 2D image $I_{2D}$ of the facial mesh $M$.  
\begin{equation}
{I_{2D}} = {F_R}(\beta ,\gamma ,\theta ,\varsigma ,\tau ,\psi ).
\end{equation}
Follow the 2D self-supervision of DECA~\cite{feng2021learning}, we use Identity, Landmark, Photometric and Shape consistency losses to form 2D loss $L_{2D}$. We can get Identity and Photometric losses through $I_{2D}$, and get landmark loss through the 2D landmark of $M$ projection. We also regularize the shape, texture, pose, expression, light and camera parameters. 

\section{Experiment}
\subsection{Implementation Setups}
\noindent\textbf{Datasets.}
To construct a dataset with 2D-3D data pairs, we download open-source facial data from FaceWarehouse~\cite{cao2013facewarehouse}, LYHM~\cite{dai2020statistical}, Stirling~\cite{feng2018evaluation} and Florence~\cite{bagdanov2011florence}, and combine them for training. We evaluate the accuracy of our monocular face reconstruction method using the NoW benchmark~\cite{sanyal2019learning} as the test data.

\noindent\textbf{Competitors.}
We select seven open-source monocular face reconstruction methods for quantitative comparison, and conduct qualitative evaluation on five of them, including Deep3DFaceRcon~\cite{deng2019accurate}, FOCUS~\cite{li2023robust}, RingNet~\cite{sanyal2019learning}, DECA~\cite{feng2021learning} and MICA~\cite{zielonka2022towards}.

\noindent\textbf{Metrics.}
In the NoW benchmark, we follow the common practice by rigidly aligning the reconstructed facial mesh with the scanned facial mesh and measuring the reconstruction accuracy using metrics including the average, median, and standard deviation of the vertices distances. We involve both non-metrical and metrical evaluation in the metrics. The scale required to provide the unit of measurements of the reconstructed mesh is calculated in non-metrical evaluation, while the scale is not calculated in metrical evaluation.

\noindent\textbf{Implementation Details.}
Our approach is implemented in PyTorch. We utilize the AdamW optimizer with a learning rate of 1e-5 and a weight decay of 2e-4. The input image size is set to 224$\times $224.  We configure the batch size as 8 and train for 160k steps. After fine-tuning the hyperparameters, we set ${\lambda_{2D}}$, ${\lambda_{3D}}$, ${\lambda_{1}}$ and ${\lambda_{2}}$ to 0.4, 0.6, 0.5 and 0.5, respectively.

\subsection{Evaluation}
\noindent\textbf{Quantitative Evaluation.}
We conduct a quantitative comparison on our proposed method and all the competitors using the NoW benchmark. Due to limited access to the entire dataset used in the MICA~\cite{zielonka2022towards}, we ensure a fair comparison by reproducing the MICA method on the available subset of the dataset and training it accordingly. Table~\ref{tab:quantitative} reports the quantitative evaluation. Compared with DECA (which only utilizes 2D supervision) and MICA (which only utilizes 3D vertices supervision), our method achieves higher reconstruction accuracy in terms of the reconstruction errors. The results of our evaluation demonstrate that the 3D encoder employed in our method effectively encodes the 3D facial mesh. This encoding process captures and represents the essential structural features of the face, thereby facilitating accurate 3D facial reconstruction. 

\begin{table}[t]
\caption{Quantitative evaluation on NoW benchmark.}
\label{tab:quantitative}
\resizebox{0.95\linewidth}{!}{ 
\begin{tabular}{c|ccc|ccc}
\hline
\multirow{2}{*}{Method}     & \multicolumn{3}{c|}{Non-Metrical (mm)$\downarrow$}  & \multicolumn{3}{c}{Metrical (mm)$\downarrow$}       \\ \cline{2-7} 
                                     & Median & Mean & Std  & Median & Mean & Std  \\ \hline
PRNet ~\cite{feng2018joint}                   & 1.50          & 1.98          & 1.88          & -             & -             & -            \\
Deep3DFaceRecon ~\cite{deng2019accurate}         & 1.23          & 1.54          & 1.29          & 2.26          & 2.90          & 2.51         \\
RingNet ~\cite{sanyal2019learning}                 & 1.21          & 1.53          & 1.31          & 1.50          & 1.98          & 1.77         \\
DECA ~\cite{feng2021learning}                    & 1.09          & 1.38          & 1.18          & 1.35          & 1.80          & 1.64         \\
FOCUS ~\cite{li2023robust}                   & 1.04          & 1.30          & 1.10          & 1.41          & 1.85          & 1.70         \\
AlbedoGAN ~\cite{rai2024towards}               & 0.98          & 1.21          & 0.99          & -             & -             & -            \\
MICA ~\cite{zielonka2022towards}                    & 0.98          & 1.21          & 1.00          & 1.25          & 1.51          & 1.30         \\
\textbf{Ours}                        & \textbf{0.93}   & \textbf{1.15} & \textbf{0.96} & \textbf{1.14}   & \textbf{1.45} & \textbf{1.23} \\ \hline
\end{tabular}
}
\end{table}

\begin{figure}[t]
    \centering
    \includegraphics[width=0.95\linewidth]{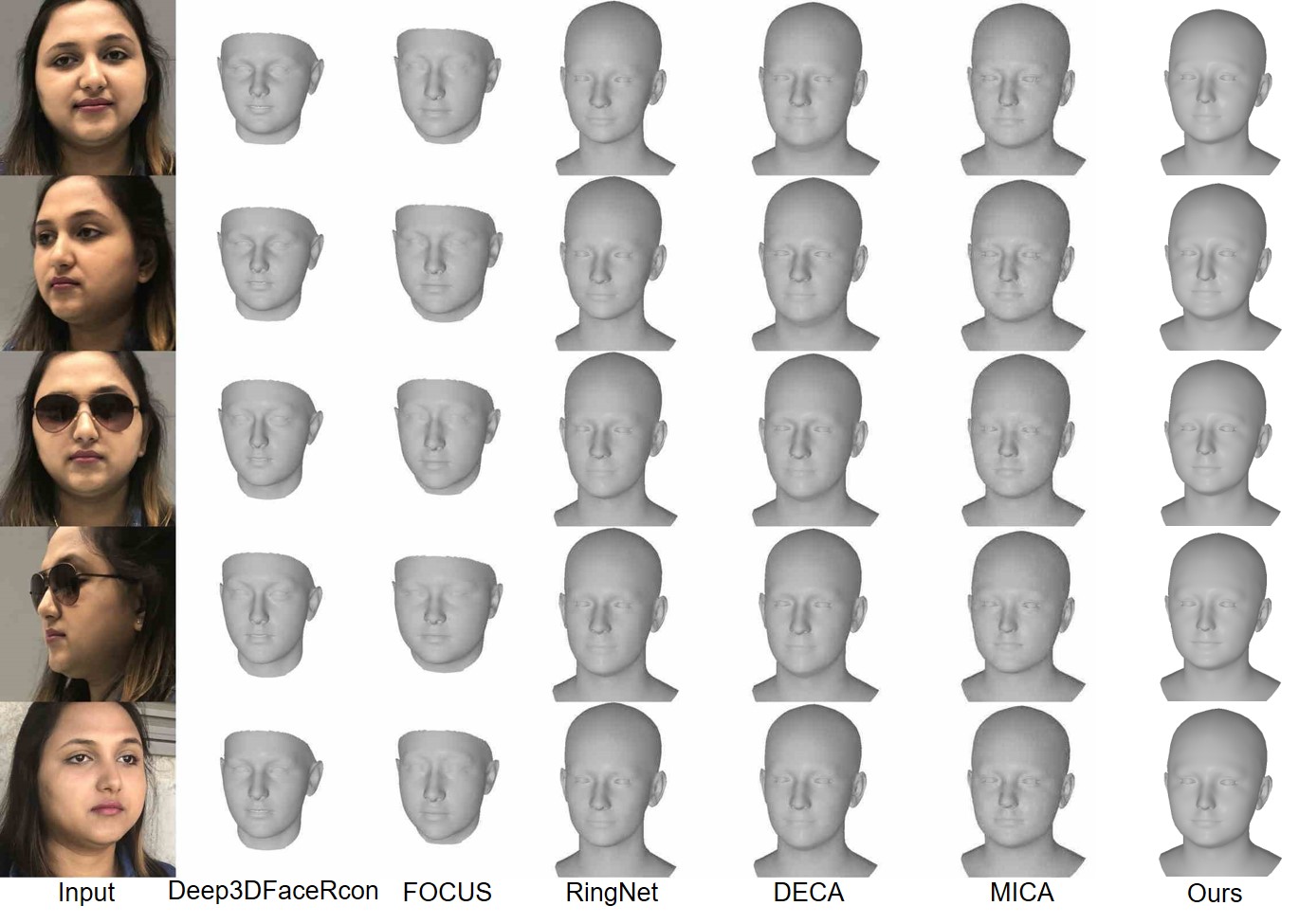}
    \vspace{-1mm}
    \caption{Visualizations of qualitative evaluation.}
    \label{fig_reconstruct}
    \vspace{-1mm}
\end{figure}

\noindent\textbf{Qualitative Evaluation.}
Assessing the quality of reconstruction results is not limited to numerical metrics alone. The performance of monocular 3D facial reconstruction methods also relies on visual presentation. 
Therefore, we conduct a visual comparison of the reconstruction performance under various conditions such as facial occlusion, different expressions, and rotations. Figure~\ref{fig_reconstruct} visualizes the qualitative evaluation. Compared with other face reconstruction methods, our method has a better reconstruction effect on facial identity features and can more realistically restore the facial shape of the input image. Please refer to more results in \url{https://github.com/Haoxin917/3DFace}.

\section{Conclusion}
In this work, we present a pipeline for monocular face reconstruction that integrates both 2D and 3D features. Our method leverages a Spectral-Based Graph Convolution Encoder, which operates on the mesh representation of the facial geometry, enabling it to capture the intrinsic 3D structural features of the face. Experimental results on the NoW benchmark demonstrate the effectiveness of our method in monocular face reconstruction. 

\small
\bibliographystyle{unsrt}
\bibliography{ref}

\end{document}